\documentclass[conference, letterpaper]{IEEEtran}
\IEEEoverridecommandlockouts
\usepackage{cite}
\usepackage{amsmath,amssymb,amsfonts}
\usepackage{algorithmic}
\usepackage{graphicx}
\usepackage{textcomp}
\usepackage[T1]{fontenc}
\usepackage{xcolor}
\usepackage{lipsum}
\usepackage{balance}
\usepackage{fancyhdr}
\usepackage{tikz}
\usepackage{lipsum}
\usepackage{subfigure}
\def\BibTeX{{\rm B\kern-.05em{\sc i\kern-.025em b}\kern-.08em
    T\kern-.1667em\lower.7ex\hbox{E}\kern-.125emX}}
\begin{document}



\title{Predicting Human Depression with Hybrid Data Acquisition utilizing Physical Activity Sensing\\ and Social Media Feeds}

\author{\IEEEauthorblockN{
\large Mohammad Helal Uddin, Sabur Baidya\\
\IEEEauthorblockA{\normalsize Department of Computer Science and Engineering, University of Louisville, KY, USA}
\normalsize {e-mail:   mohammad.helaluddin@louisville.edu, sabur.baidya@louisville.edu}
\vspace{-6mm}
}}

\maketitle

\pagestyle{fancy}
\thispagestyle{fancy}
\renewcommand{\headrulewidth}{0pt}  
 
\fancyhf{}
 
\fancyhead[C]{
    \small This article has been accepted for publication in the International Conference on Pervasive Computing and Communications (IEEE PerCom 2025) Workshop. 
}
 
\fancyfoot[C]{
    \begin{tikzpicture}[remember picture, overlay]
        \node[anchor=south,yshift=10pt] at (current page.south) {
            \parbox{\textwidth}{
                \centering
                \footnotesize
                \textcopyright~2025 IEEE. Personal use of this material is permitted. Permission from IEEE must be obtained for all other uses, in any current or future media, including reprinting/republishing this material for advertising or promotional purposes, creating new collective works, for resale or redistribution to servers or lists, or reuse of any copyrighted component of this work in other works.
            }
        };
    \end{tikzpicture}
}

\begin{abstract}
Mental disorders including depression, anxiety, and other neurological disorders, pose a significant global challenge, particularly among individuals exhibiting social avoidance tendencies. This study proposes a hybrid approach 
by leveraging smartphone sensor data measuring daily physical activities and analyzing their social media (Twitter) interactions for evaluating an individual's depression level . 
Using CNN-based deep learning models and Naive Bayes classification, we identify human physical activities accurately and also classify the user sentiments.
A total of 33 participants were recruited for data acquisition 
and nine relevant features were extracted from the physical activities and analyzed with their weekly depression scores, evaluated using the Geriatric Depression Scale (GDS) questionnaire. 
Out of the nine features, six are derived from physical activities, achieving an activity recognition accuracy of 95\%, while three features stemmed from sentiment analysis of Twitter activities, yielding a sentiment analysis accuracy of 95.6\%. Notably, several physical activity features exhibited significant correlations with the severity of depression symptoms. For classifying the depression severity, a support vector machine (SVM) based algorithm is employed that demonstrated very high accuracy of 94\%, outperforming the alternative models, e.g., the multilayer perceptron (MLP) and k-nearest neighbor. 
It's a simple approach yet highly effective in the long run for monitoring depression without breaching personal privacy. 
\end{abstract}


\vspace{-2mm}
\section{Introduction}
Mental wellness is a crucial aspect of our overall wellbeing, and the World Health Organization (WHO) defines it as a state in which individuals can achieve their potential, handle normal stresses in life, be productive, and contribute to their communities~\cite{b1}. Our mental wellness is influenced by various factors, such as genetic, sociocultural, economic, political, and environmental perspectives. In recent years, mental health issues have become a significant public health concern, 
causing devastating effects on patients and their families~\cite{b2} \cite{b3}. According to the WHO, depressive disorders are among the most common mental illnesses~\cite{b4}. Symptoms of these disorders include heartbreak, loss of interest and joy, feelings of guilt or low self-esteem, sleep disorders or cravings, feelings of weakness, and poor concentration~\cite{b4}. In Bangladesh, where this study is conducted, more than 6 million individuals are affected~\cite{b5}, and according to the research, it's claimed that depression is going to be the 2nd most alarming sickness in the world by 2024, after the ischemic heart disease \cite{b6}.\\
Effective treatment strategies,e.g., interpersonal psychotherapy, cognitive social therapy , and others have been found to cure most psychological illnesses \cite{b7}. However, the efficacy of these treatment methods is crucial to ensure proper treatment. Although various self-report based methods have been evaluated for identifying the nature of depression \cite{b10}, recognizing specific symptoms on one's own can be difficult, and relying solely on self-report methods can introduce methodological biases, social desirability biases, and memory recall biases~\cite{b8} - \cite{b13}. Furthermore, the high cost of medical care means that people in underdeveloped and developing countries often receive inadequate or no assistance at all~\cite{b11}. Depressed patients also avoid inpatient treatments as they are unable to continue long-term supervision for their diagnosis and treatment or try to manage the challenges independently~\cite{b12}. \\
Now in recent times, the use of smartphones has become an essential part of people's daily lives. It is expected that the number of smartphone users will increase from 5.7 billion in 2020 to 7.7 billion in 2028 \cite{b14}. In Bangladesh, where the current study was conducted, there were an estimated 105 million smartphone users \cite{b14}. This accounts for approximately 61\% of the total population. With the advancement of smartphone hardware, high-level programming, and sensing capabilities, they can now be used to monitor an individual's wellness, daily activities, location, and provide guidance based on environmental factors such as weather forecasting and climate prediction. \\
In recent years, researchers have focused on using sensor-based services to monitor individuals who live independently, as these technologies can extract characteristics of behavior that can be used to identify symptoms of depression~\cite{b15} - \cite{b20}. Smartphone sensors can detect symptoms such as anhedonia, lassitude, and psychomotor retardation, which are associated with physical activities (such as running, walking, and sleeping)~\cite{b21}, changes in behavior (e.g., cell phone call sequences~\cite{b22}, circadian movement~\cite{b23}), and social interactions~\cite{b24}. One study examined the physical activities of mentally ill patients using smartphones and found a correlation between changes in psychological state and activities such as running, walking, and sleeping~\cite{b22}. Another study developed a sensor-based monitoring method for detecting depression severity in the elderly using daily physical activities in their living space~\cite{b25}. Passive data collected by mobile sensors is potentially more powerful than traditional self-reports because it overcomes methodological biases, social desirability biases, and memory recall biases. The data collected by mobile sensors can be used to manage depression and monitor individuals at risk, without incurring traditional treatment costs~\cite{b26}. \\
Social media platforms such as Twitter and Facebook have become essential tools for individuals with depression to share their experiences and seek support and advice \cite{b27} - \cite{b29}. The widespread availability of smartphones has made it easier for people to access these platforms at any time and from any location. Twitter is particularly useful in this regard, as it provides a wealth of social data that can be used to identify key features of mental disorders \cite{b30}. In the field of health, social media has become an important source of information, allowing for the detection and prediction of mental disorders and serving as a tool for mental health monitoring and surveillance \cite{b31} - \cite{b34}. The use of natural language processing (NLP) and machine learning (ML) technologies has also proven effective in detecting early symptoms of depression through the analysis of online data \cite{b35}-\cite{b37}. Twitter, with over 330 million users worldwide, is one of the most popular social networking sites \cite{b38}\cite{b39}. By analyzing the language and social behavior in tweets and posts, researchers can gain insight into personality traits, lifestyles, and mental health conditions \cite{b40}-\cite{b44}. 

In this work, we have proposed a new hybrid data acquisition model with social network information and a smartphone app that uses the sensor data, followed by a CNN algorithm with depth-wise separable convolution to recognize physical activities and the naïve-Bayes classification algorithm to determine tweet sentiment (negative, neutral, or positive). 
The summary of our contributions are as follows: 
\vspace{-1mm}
\begin{itemize}
    \item We proposed a novel hybrid data acquisition model that includes physical data from smartphone sensors and social media data on the same individuals under the study. To the best of our knowledge this is the first hybrid data acquisition with sensor data and social media data in context of assessing and predicting human depression.
    \item We developed an end-to-end analytics framework using a composite of deep learning and machine learning models for the classification of activities, sentiment analysis and finally, the prediction of human depression.
    \item We conducted extensive experiments with cross validation with our proposed framework on the collected data from human participants, and our results showed that the SVM method had the highest accuracy rate of 94\% for severe depression, performing better than other classification models such as MLP and k-nearest neighbor. The proposed method outperforms the existing depression classification techniques \cite{b49} by     $\sim$8\% increase in accuracy.
\end{itemize}

\section{Data Collection and Preprocessing}
Our study involved collecting data from three different sources: accelerometer, GDS (Geriatric Depression Scale), and Twitter. We gathered this data using a hybrid data acquisition model, which we implemented at a medical institute in Bangladesh. Our study involved 33 participants, and we collected data using a specific procedure that we detail in the following section with the help of Figure 1. We describe the dataset in two sections, namely Dataset I and Dataset II.
\begin{figure}[htbp]
\vspace{-2mm}
\centerline{\includegraphics[scale=0.25]{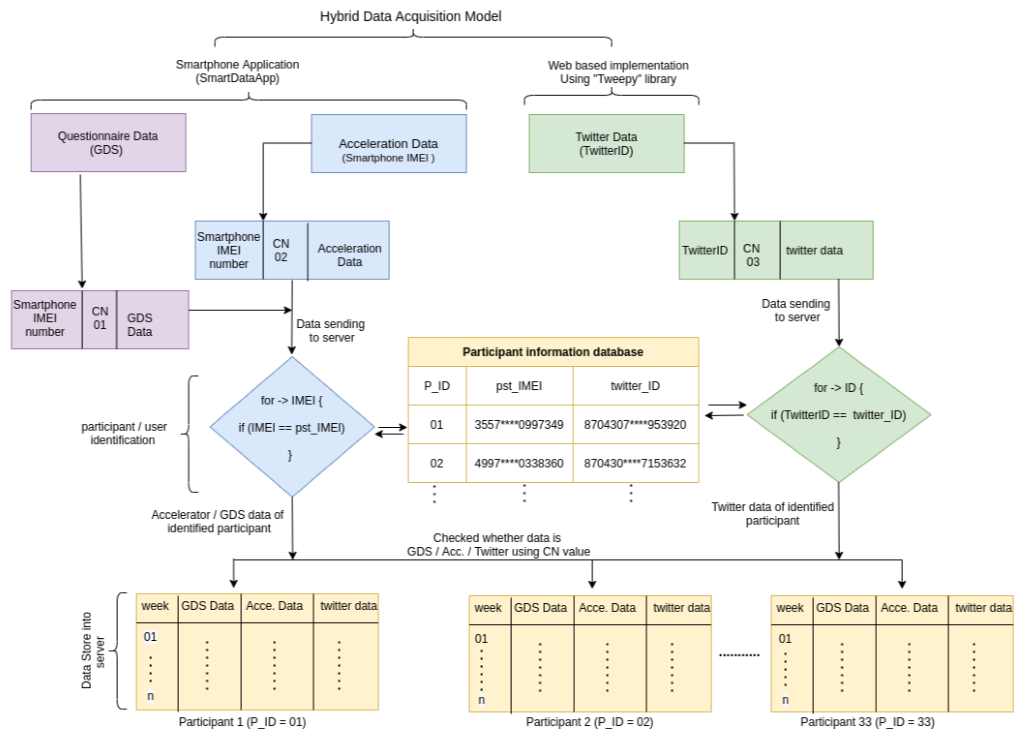}}
\vspace{-2mm}
\caption{Hybrid data acquisition model.}
\label{fig 1}
\vspace{-4mm}
\end{figure}
\subsection{Hybrid Data Acquisition Model}

In this work, we introduced a new automated data acquisition model to collect three-dimensional data from each participant. To ensure accurate analysis, it was crucial to synchronize all three dimensions of data. We used an Android-based mobile application called "SmartDataApp" to collect questionnaire data (GDS) and smartphone acceleration data. This application was specifically designed to store one week of acceleration data in its internal data storage system. After one week, the application automatically sent the acceleration data to the server using an API. This approach ensured that we had a reliable and consistent data set to work with.

\begin{figure}
\vspace{2mm}
\hfill
\subfigure{\includegraphics[width=2cm]{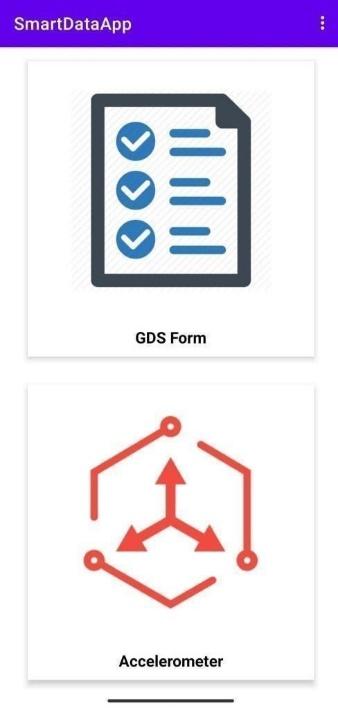}}
\hfill
\subfigure{\includegraphics[width=2cm]{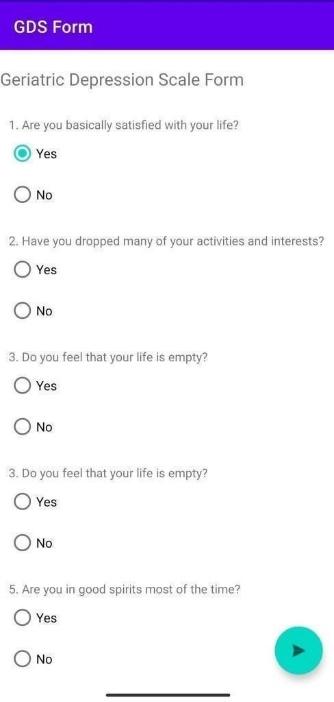}}
\hfill
\subfigure{\includegraphics[width=2cm]{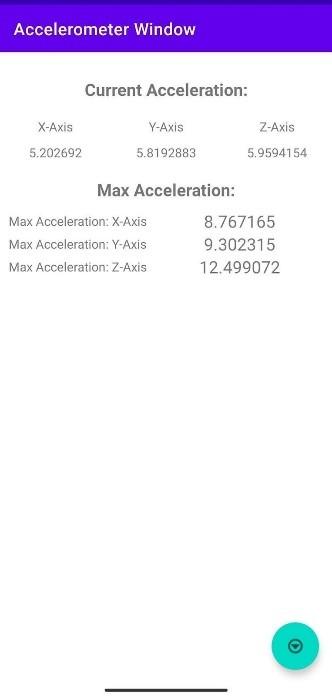}}
\hfill
\subfigure{\includegraphics[width=2cm]{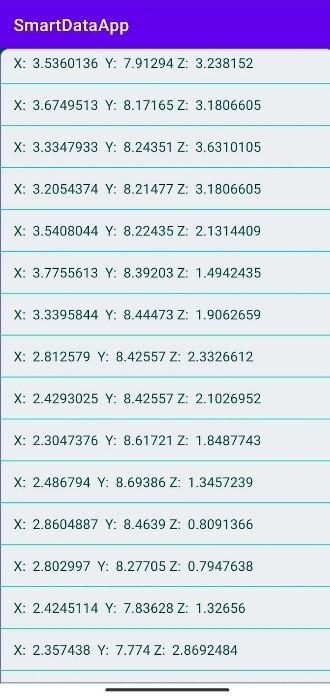}}
\hfill
\caption{SmartDataApp application, (from left) application main window, GDS form window, real-time accelerometer data visualization, and data are saved into internal storage, Accelerometer data preview from internal application storage.}
\label{fig 2}
\vspace{-4mm}
\end{figure}
For Twitter data, an API based (using tweepy library) system has been introduced. Participants' identification was made using smartphone IMEI number (while data was coming from SmartDataApp) and twitter\_id (while data was coming from Twitter). All participants' primary data information (IMEI number, Twitter id) were pre-stored in the database before the experiment was placed. SmartDataApp application was installed for each participant's smartphone before the experiment started; therefore, the smartphone IMEI number was used as a unique identification. Twitter has a unique id number for each of its users, so there is no possibility of twitter id duplication. While the data comes to the server, the server can identify the participant from the IMEI number and twitter\_id. Once data is identified against a specific participant, the server checks while the data is GDS data/acceleration data/ Twitter data, using "CN" parameter value. CN=01 means GDS data; CN=02 means acceleration data; CN=03 means Twitter data.

\subsection{Dataset I}
We collected data from 33 participants, consisting of 19 males and 14 females, all over 18 years old with a mean age of 24 years and a standard deviation of ±5. The data collection period was from April 2019 to June 2020, resulting in 60 weeks of activity data. However, we only considered 55 weeks of data for analysis. All participants were volunteers from a medical institute in Dhaka, Bangladesh, who agreed to use the SmartDataApp to collect their data. We collected physical activity data using the accelerometer in participants' smartphones via the SmartDataApp. Throughout the study, participants were encouraged to keep and carry their smartphones with them at all times. Figure \ref{fig 2} (most right) displays the raw data collected from the mobile sensor's X-axis, Y-axis, and Z-axis via the data collector application.

To assess the mental health condition of the participants, we used the Geriatric Depression Scale (GDS)\cite{b45}, which has been validated in previous research \cite{b46}. We categorized the depression level based on the GDS score into three categories: absence of depression (GDS < 5), moderate depression/mild depression (5 $\le$ GDS $\le$ 9), and severe depression (GDS > 9) \cite{b46}. Using the SmartDataApp, we collected the GDS score from the participants every week. The first week's depression scores are presented in Table 1.

\begin{table}[htbp]
\vspace{-2mm}
    \centering
    \caption{Depression scores distribution}
    \label{tab:example}
    \begin{tabular}{|p{22mm}|p{16mm}|p{16mm}|p{14mm}|}
        \hline
          & GDS<5 & 5 $\le$ GDS $\le$ 9 & GDS > 9 \\
        \hline
        Participants & 23 & 7 & 3 \\
        \hline
        Depression level & AB & MD & SD \\
        \hline
        Gender & 9 females, 14 males & 4 females, 3 males & 1 female, 2 males \\
        \hline
        Age +/-SD (years) & 24 (+/- 4) & 25 (+/- 5) & 24 (+/-4) \\
        \hline
    \end{tabular}
    \vspace{-2mm}
\end{table}

\subsection{Dataset II}
In this study, we also collected Twitter data from the same 33 participants. To collect the data, we used the "tweepy" Twitter API \cite{b47} and applied a specific filter using specific keywords. The filter was designed to capture data by searching for keywords related to depression, anxiety, sadness, and mental health. Specifically, we used the following filter: \textit{stream.filter(track=['Depression', 'Anxiety', 'Sad', 'mental health'])}. This filter allowed us to sort out the depression-related tweets from other tweets. In total, we collected 3500 tweets from the 33 participants. 

\subsection{Data Preprocessing}
The study collected three types of data from each individual: accelerometer data, Twitter data, and Geriatric Depression Scale (GDS) scores. The aim of data set I was to classify physical activities, such as \textit {sitting, walking, standing,} and \textit {jogging}. Unprocessed acceleration data was collected at a 20 Hz sample rate. To remove noise peaks resulting from acceleration, a third-order low-pass Butterworth filter was used. The data was then filtered and subtracted to separate the gravitational component from the body acceleration data. A third-order low-pass Butterworth filter with a cutoff frequency of 0.3Hz was used to separate the gravitational component \cite{b48}\cite{b49}. For data set II, Hadoop was used to preprocess Twitter data collected using the "tweepy" Twitter API. Each instance of raw Twitter data contains many pieces of information, but for this study, only the "id" and "text" were used. The "id" object contains an 18-digit ID number, and the "text" object contains the tweet itself. The "text" column was used for the sentiment analysis of participants.

\section{Method}
\begin{figure}
\vspace{2mm}
\hfill
\subfigure{\includegraphics[scale=0.32]{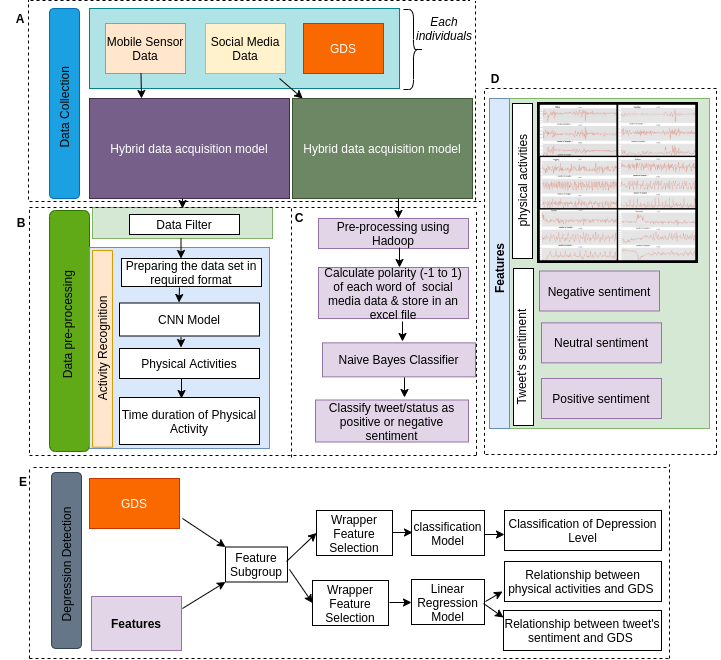}}
\hfill
\subfigure{\includegraphics[scale=0.32]{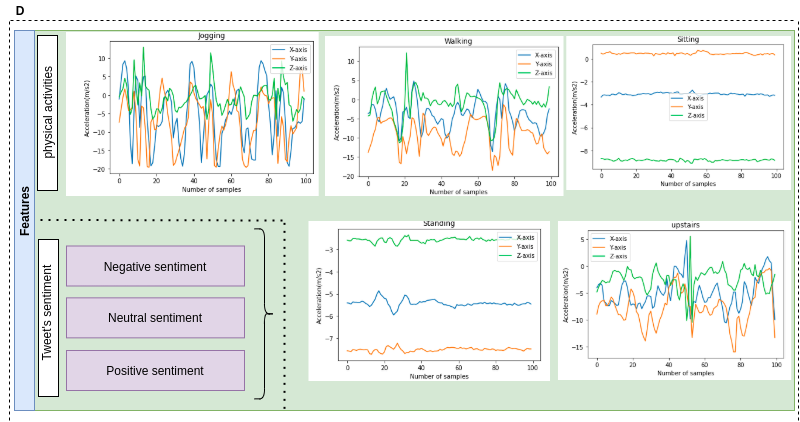}}
\hfill
\vspace{-6mm}
\caption{Work flow chart consisting of (A) data collection, (B) mobile data preprocessing activity recognition, (C) Twitter data preprocessing sentiment analysis, (D) Features, (E) Depression detection.}
\label{fig 3}
\vspace{-4mm}
\end{figure}

This study aimed to develop a model that could classify depression severity levels based on smartphone sensor information and social media activity. The focus of the study was to analyze physical and social media behaviors to assess the depression severity level of individuals. The system workflow consisted of four stages, as shown in Figure \ref{fig 3}. Three-dimensional data were collected (Figure \ref{fig 3}A), and the acceleration data was filtered to ensure uniform sampling and reduce noise (Figure \ref{fig 3}B). The filtered data features were then used to detect individuals' activity patterns in the activity recognition stage. Twitter data were collected using the "tweepy" API, and the data was filtered to extract the necessary information (Figure \ref{fig 3}C). Features from activities and sentiments from tweets were then combined (Figure \ref{fig 3}D), a regression and classification model was applied to the collected data to predict depression scores and levels (Figure \ref{fig 3}E).

\subsection{Physical Activity Recognition}\label{3A}
The CNN model architecture consists of an input layer, output layer, and in between them hidden layers. The hidden layers usually comprise with the convolutional layers, ReLU layers, pooling layers, and fully connected layers. Figure \ref{fig 3}D is the visualization of the component of the accelerometer for different physical activities. The visualization of \ref{fig 3}D has plotted the 9sec signal for physical activities.

\begin{figure}[!t]
\vspace{2mm}
\centerline{\includegraphics[width=0.9\linewidth]{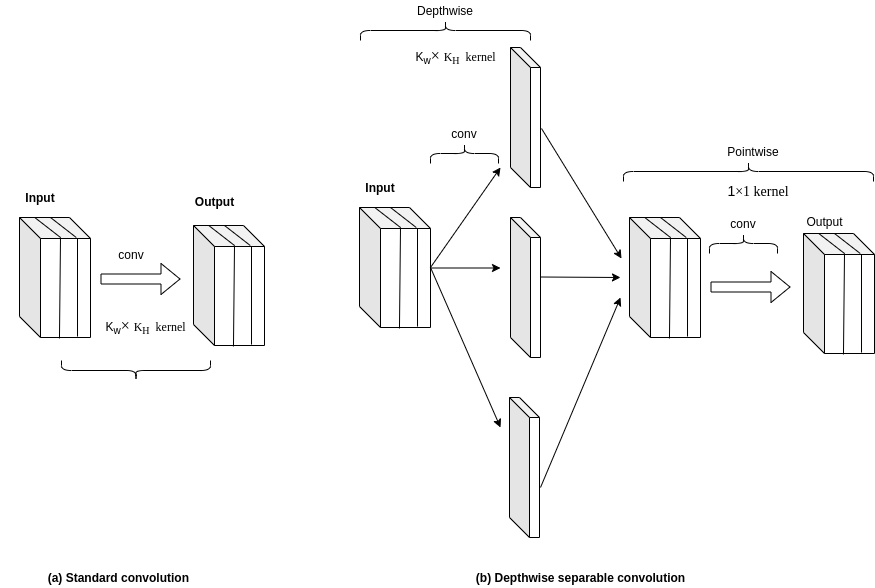}}
\caption{Standard convolution and Depthwise convolution}
\label{fig 4}
\vspace{-4mm}
\end{figure}

This work employs depthwise separable convolutions, which reduce the computational cost and weight parameters. In deep learning, the regular convolution is commonly referred to as standard convolution, and its basic operations are depicted in Figure \ref{fig 4}a. Each output feature map of a standard convolution layer with $C_{in}$ input channels and $C_{out}$ output channels contains the sum of $C_{in}$ input function maps convoluted by the $C_{in}$ kernels. The number of weights is in the standard convolution:
\vspace{-2mm}
\begin{equation}\label{eq.1}
    W_{std} = C_{in} \times k_w \times k_H \times C_{out}
\end{equation}
Where $k_W$ $\times$ $k_H$ represents kernel size. the computational cost of the output feature maps with size $f_W$ $\times$ $f_H$, is:
\begin{equation}\label{eq.2}
    CC_{std} = C_{in} \times k_w \times k_H \times C_{out} \times f_W \times f_H
\end{equation}

Figure \ref{fig 4}b demonstrates the work of depthwise convolution and depthwise separable convolution. The number of weights of a depthwise convolution is:
\begin{equation}\label{eq.3}
    W_{dw} = k_{w} \times k_H \times C_{out}
\end{equation}
 In a depthwise convolution layer, the computational cost is:
 \begin{equation}\label{eq.4}
    CC_{stw} = k_w \times k_H \times C_{out} \times f_W \times f_H
\end{equation}

The model for this work includes a layer with max pooling, which is then followed by another convolution layer. After this, the model includes a fully connected layer that is connected to the softmax layer. Both the convolution and max-pool layers are 1D or temporal.

\begin{figure}[htbp]
\centerline{\includegraphics[width=0.99\linewidth]{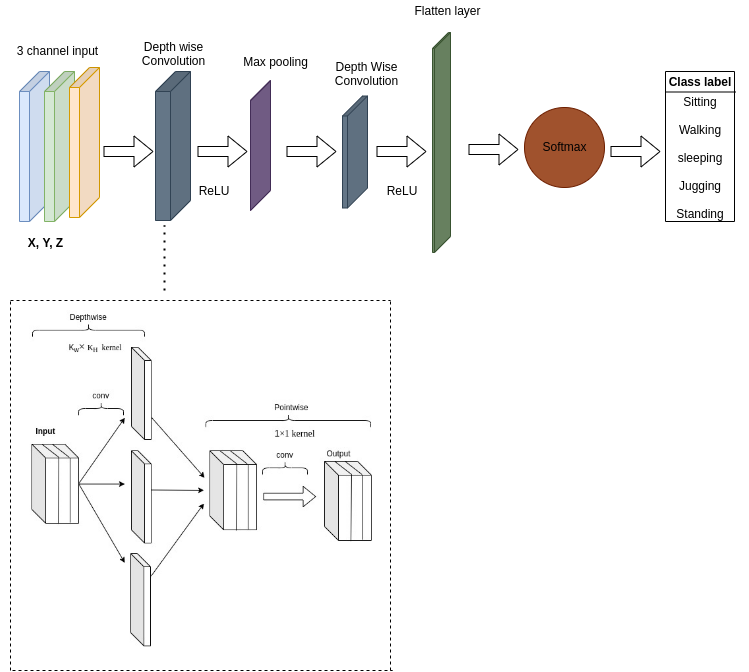}}
\caption{CNN Model with depthwise convolution}

\label{fig 5}
\vspace{-6mm}
\end{figure}
\vspace{10pt}

The first convolution layer of the model uses a filter size and depth of 60. The pooling layer that follows has a filter size of 20 and a stride of 2. The next convolution layer applies a filter size of 6 after taking input from the max-pooling layer. Its depth is one-tenth of the depth of the max-pooling layer. The output from this layer is flattened and fed as input to the fully connected layer, which has 1000 neurons. A \textit {tanh} function is applied to introduce non-linearity in this layer. The SoftMax layer determines the output probabilities of the class labels. To minimize the negative log-likelihood cost function, a stochastic gradient descent optimizer is used. 

\begin{table}[htbp]
    \centering
    \caption{Data representation of participant 1 and 2}
    \label{table 2}
    \begin{tabular}{|p{10mm}|p{11mm}|p{13mm}|p{13mm}|p{8mm}|p{9mm}|}
        \hline
        Particip-ant's ID & X-accel & Y-accel & Z-accel & Activity & Time (ns) \\
        \hline
        01 & 0.46309182 & -2.4108016 & 2.4108016 & Jogging & 4910989 234900  \\
        \hline
        01 & 2.4925237 & 8.730643 & -1.4573772 & Walking & 4939552 229300 \\
        \hline
        02 & 0.8036005 & 2.4108016 & 3.706926 & Jogging & 58289822 191000 \\
        \hline
        02 & 2.0294318 & 9.956474 & 0.7218784 & Walking  & 58121982 286000 \\
        \hline
    \end{tabular}
    \vspace{-4mm}
\end{table}

In Table \ref{table 2}, from columns 1st to 6th represents the participant's id, X- acceleration, Y- acceleration, Z-acceleration, physical activity, and time (ms). Time is represented as nano-seconds.

To analyze the social media data, we use the following terminologies:\\
Polarity: The connotation of each word, i.e., positive (+1), negative (-1) or neutral (0) in a sentence. This gives a polarity score of a sentence as positive (>0), negative (<0) or neutral (=0).\\
Sentiment: Sentiment analysis involves examining tweets to determine if they are positive or negative. Imagine each tweet as a mix of words, with some being happy words and others sad. The overall sentiment, or mood, of the tweet is calculated using a formula called $TS_p$ (equation 5), which weighs these words. Just like how a scale measures weight,$TS_p$ measures whether a tweet leans more towards happiness or sadness, giving us insight into an individual's opinion on different tweets.

In equation \ref{eq.5}, $TS_p$ represents the polarity score of a tweet, where $w_1$+$w_2$+$w_3$+ ….+$w_n$ means each word of the tweet, and $p_1$+$p_2$+$p_3$+ …. +$p_n$ represents the polarity of the words.
 \begin{equation}\label{eq.5}
    TS_p = \sum{w_1p_1 + w_2p_2 + w_3p_3 + ..... + w_np_n}
\end{equation}
The polarity of each word can be either -1, 0 or +1 and the polarity score of each tweet will be greater than zero, equal to zero or less than zero. Consequently, each tweet will receive a polarized score, where a negative score indicates a negative tweet, and a positive score indicates a positive tweet. Words like articles and prepositions are considered neutral (with a polarity of 0). If a tweet's polarized score is 0, that indicates the tweet is neither negative nor positive; it's a neutral tweet. The table below shows examples of tweets with positive, negative, neutral polarizations.

\begin{table}[htbp]
    \centering
    \caption{Examples of text classification}
    \label{table 3}
    \begin{tabular}{|p{23mm}|p{9mm}|p{10mm}|p{8mm}|p{15mm}|}
        \hline
        Sentence & Positive Words & Negative Words & Neutral words & Polarization \\
        \hline
        The joke is so funny, I am laughing like dying & joke, funny, laughing, like & Dying & The, is, so, I & +1 (positive) \\
        \hline
        I am so depressed I will suicide oneday & oneday & Depressed, suicide & I, am, will, so & -1 (negative) \\
        \hline
        Someday I will die & Someday & Die & I,will & 0 (neutral) \\
        \hline

    \end{tabular}
\end{table}

This procedure classifies tweets as either positive or negative. The entire dataset is processed using the same method, and the classified tweets are stored in an Excel file. This file is then used as input to the classifier model to determine the sentiment expressed in the tweets. We have used Naïve Bayes classification model \cite{b50} for this work which works as follows:
The vector of dependent feature ($v_1$,...,$v_n$), class is $S_c$.
 \begin{equation}\label{eq.6}
    P(S_c | v_1,...,v_n) = \frac{P(S_c)P(v_1,...,v_n | S_c)}{P(v_1,...,v_n)}
\end{equation}
 \begin{equation}\label{eq.7}
    P( v_i| S_c,v_1,...,v_n) = P(v_i | S_c)
\end{equation}
 \begin{equation}\label{eq.8}
    P(S_c | v_1,...,v_n) = \frac{P(S_c) \prod_{i}^{n} P(v_i | S_c)}{P(v_1,...,v_n)}
\end{equation}
where i=1
 \begin{equation}\label{eq.9}
    P(S_c | v_1,...,v_n) \propto {P(S_c) \prod_{i}^{n} P(v_i | S_c)}
\end{equation}
\begin{equation}\label{eq.10}
    \Bar{y} = {argmax}_c {P(S_c) \prod_{i}^{n} P(v_i | S_c)}
\end{equation}
\begin{equation}\label{eq.11}
    \Bar{y} = {argmax}_c {(\ln{P(S_c)} + \sum_{i}^{n} \ln{P(v_i | S_c)})}
    \vspace{-1mm}
\end{equation}
For each class $S_c$, multinomial distribution is parametrized by the vector $\alpha_c$= $\alpha_1$ ... $\alpha_n$.
\begin{equation}\label{eq.12}
    \alpha_{ci} = \frac{N_{ci} + \gamma}{N_c + \gamma_n}
\end{equation}
$N_{ci}$ represents the number of times  \textit{i}(feature) appears in a particular sample of class c in the time of training period. $\gamma$ is the smoothing priors, when $\gamma$=1 represents Laplace smoothing and $\gamma$ < 1 represents \textit{Lidstone} smoothing.
\begin{equation}\label{eq.13}
    \Bar{y} = {argmax}_c {(\ln{P(S_c)} + \sum_{i}^{n} \ln{\frac{N_{ci} + \gamma}{N_c + \gamma_n}})}
\end{equation}

\subsection{Feature Subgroup}
A set of nine features were derived from the classification of participants' physical activities and Twitter text. The features "sitting", "walking", "standing", "jogging", "going upstairs", and "going downstairs" quantify the level of physical activity for each participant. Additionally, the "negative", "neutral", and "positive" sentiment features determined the participant's sentiment based on their Twitter text. 

\begin{table}[htbp]
\vspace{2mm}
    \centering
    \caption{Features of participants’ activity.}
    \label{table 4}
    \begin{tabular}{|c|c|}
        \hline
        Features & Abbr. \\
        \hline
        ST & Sitting time (Physical activity) \\
        \hline
        WT & Walking time (Physical activity) \\
        \hline
        StT & Standing Time (Physical activity) \\
        \hline
        JoT & Jogging Time (Physical activity) \\
        \hline
        UpT & Going Upstairs time (Physical activity) \\
        \hline
        DownT & Going Downstairs time (Physical activity) \\
        \hline
        NeS & Negative Sentiment (Twitter data analysis) \\
        \hline
        NuS & Neutral Sentiment (Twitter data analysis) \\
        \hline
        PoS & Positive Sentiment (Twitter data analysis) \\
        \hline
    \end{tabular}
    \vspace{2mm}
\end{table}

\vspace{2mm}
This study has found that there is a relationship between the features of working and non-working days with participants' depression scores. As a result, the features were divided into three subgroups: i) weekdays (Saturday to Friday, n=7), ii) working days (Sunday to Thursday, n=5), and iii) weekends (Friday-Saturday, n=2). 
\begin{equation}\label{eq.14}
    FS_n = {\sum_{i}^{n} F_i}
\end{equation}
Here, $FS_n$ means subgroup of features, i = 1,2,3 .... n, $F_i$ means $i_{th}$ day features.

\subsection{Feature Selection}

The wrapper feature selection method \cite{b56}\cite{b57} was utilized in this study for selecting features. The chosen features were evaluated using regression and classification models. The evaluation process was carried out through tenfold cross-validation of the sample. The features for the regression model were selected based on the RMSD (root mean square deviation) of the GDS score assessment. \\
The GDS score was assessed at the outset of the study, and individual depression severity was determined. The current study observed a correlation between each extracted feature and depression symptoms, which can differentiate whether a participant is experiencing depression or not. To estimate the GDS score, a linear regression model \cite{b51} was applied to the extracted features. 
 \begin{equation}\label{eq.15}
    GDS'_i = \sum{\beta_{0} + \beta_1x_1 + \beta_2x_2 + ..... + \beta_nx_n}
\end{equation}

Here, $GDS'_i$ is the projected GDS points for $i$ participants, $x_n$ is the $n_{th}$ numbers of features. $\beta_0+ \beta_1+ \beta_2+…...+ \beta_n$ are the coefficients of the linear regression model. The lasso regularization \cite{b52} was applied to limit the error of high dimensional features between the projected score and the genuine GDS score. This helps to prevent the regression model coefficient from getting excessively too large. Since all the features had solid correlations, The Lasso regularization has shown a perfect fit to this model.

\subsection{Level Classification of Depression Severity}

There were three levels (Table 1) for depression classification, and for observing the classification execution, three very well-known classification algorithms Support Vector Machine (SVM) \cite{b53}, K-nearest neighbor (KNN) \cite{b54}, Multilayer perceptrons (MLP)\cite{b55} were considered.

\subsection{Evaluation and Validation}
For evaluating both of the model linear regression and classification, LOOCV(leave-one-out cross-validation) method was used. During the validation process, a single participant labeled(weekly) data was used for testing where the rest of the participants data was used for training . It’s a repeated cycle for each participant to accomplish the best outcome. Corresponding to recognizing depression levels, classes were not similarly conveyed, the number of data samples from a single class was evidently greater than the number of data samples from various classes., hence by doing the subsampling over-represented, the training samples were synthesized which was applied to the LOOCV method. Calculating the error difference between the projected score ($GDS_i$) and real score (GDS), RMSD (root mean square deviations) was applied. The model performance was determined by the RMSD, the calculated equation was as bellows:

 \begin{equation}\label{eq.16}
    RMSD = \sqrt{\frac{1}{N} \sum_{i}^{N}{(GDS - GDS'_i)^{2}}}
\end{equation}

\vspace{2mm}
\section{Results}

\subsection{Sentiment Analysis From Twitter Data}
\begin{figure}[htbp]
\vspace{2mm}
\centerline{\includegraphics[width=0.95\linewidth, height = 6.5cm]{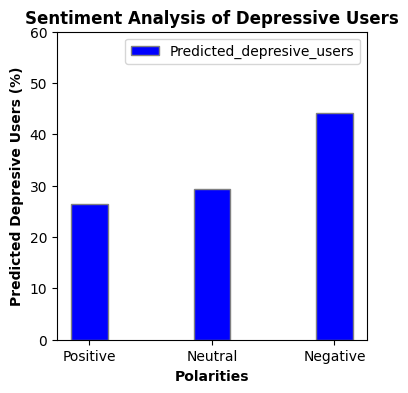}}
\vspace{-2mm}
\caption{Predictive depressive users in each polarities.(X-axis represents Polarity, Y-axis represents Frequency (\%)).}
\label{fig 6}
\vspace{-4mm}
\end{figure}

This study conducted sentiment analysis on tweets in two stages: statistical analysis for data visualization and Naïve Bayes classification to classify tweets. In terms of finding the sentiment, this study performed a polarity analysis of the tweets. To perform the polarity analysis, the Twitter data set was divided into three domains, "Positive polarity", "Neutral polarity", and "Negative polarity". The sentiments of the tweets (positive, negative, neutral) are calculated using equation (5). This study shows that figure \ref{fig 6} and Table \ref{table 5}, in the positive polarity domain; 26.5\% (9 out of 33) participants are predicted depressed users where positive sentiment tweets are 67\% and negative sentiment tweets are 12\%. In the neutral polarity domain, there are 29.4\% (10 out of 33) participants are predicted depressed users, where positive sentiment tweets are 13\% and negative sentiment tweets are 8\%.

\begin{table}[h]
\centering
\caption{All participants' tweet’s polarity indication according to Positive, Neutral, and Negative.}
\label{table 5}
\begin{tabular}{|c|c|c|}
\hline
Polarities & +ve sentiment tweets(\%) & -ve sentiment tweets(\%) \\ \hline
Positive & 67 & 12 \\ \hline
Neutral & 13 & 8 \\ \hline
Negative & 20 & 80 \\ \hline
\end{tabular}
\end{table}

The most notable findings emanated from the "Negative Polarity" domain, where a substantial 44.12\% of participants (14 out of 33) as shown in figure \ref{fig 6} and Table \ref{table 5}, exhibited depressive tendencies. Table \ref{table 5} shows that negative sentiment tweets dominated this domain at 80\%, while positive sentiment tweets comprised only 20\%. This disparity underscores a pronounced correlation between negative sentiment expression on Twitter and the likelihood of users experiencing depression. The high prevalence of depressed users in this domain accentuates the potential role of negative sentiment as an indicator or influencer of mental health issues among the participants of Twitter users. This study underscores the intricate interplay between sentiment expressed in tweets and the mental well-being of users. The finding suggested that the negative sentiment tweets appears to correlate more strongly with the possible depression. \\

\begin{figure}[htbp]
\centerline{\includegraphics[scale=0.42]{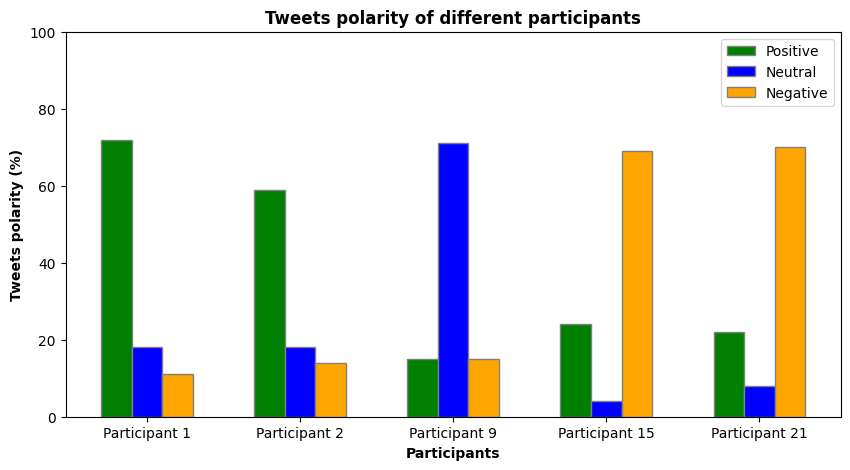}}
\vspace{-4mm}
\caption{Participant's sentiment visualization.}
\label{fig 7}
\vspace{-2mm}
\end{figure}

In figure \ref{fig 7}, we have randomly chosen 5 participants out of 33 to visualize their sentiment based on tweet polarity. In figure \ref{fig 7}, Participants 1 and 2's most of tweets are positive (71.42\% and 59.09\%), where neutral tweets are 17.85\%, and negative tweets are 10.71\% and 13.63\%, which indicates positive sentiment. In a similar way, in Participants 15 and 21's most of the tweets are negative (68.57\% and 70\%), where neutral tweets are 4.76\%, and positive tweets are 24.76\% and 22.22\%, which indicates negative sentiment. According to the statistical analysis in the previous section (figure 6), Participants 1 and 2, exhibiting predominantly positive sentiments, might suggest a correlation between optimistic online expression and positive mental well-being. It's conceivable that individuals with a generally positive outlook in their tweets may experience lower levels of depressive tendencies.\\

Participant 9 is the only random participant among these 5 participants whose sentiment is neutral, where the positive tweets percentage is 15.83\%, the negative tweet's percentage is 14.54\%, and the neutral tweet's percentage is 70.83\%. The correlation between participants and tweet polarity in the presented analysis unveils intriguing patterns that illuminate the nuanced dynamics of sentiment within this select group.  \\
For classifying the tweets, we performed the Naïve Bayes classification algorithm. We have fed our pre-processed data into this classifier. We have got 95.6\% accuracy while classifying tweets into negative, positive, and neutral. 



\subsection{Activity Recognition}

Activity recognition theoretical methodology is explained in section 2.4.  We have initialized our batch size as 10 and run the test for 12 epochs. We have run the test for a few times and observed the output. We have found a quite consistent performance of the model. The CNN model accuracy has obtained around 98\% while identifying activity. 


\subsection{GDS Score Prediction}
The study categorized 9 features related to depression score into 3 subgroups based on the participant's activities. Each feature was separately evaluated using a linear regression model. The depression levels were classified into three groups, namely absence (GDS < 5), mild/moderate (5 $\le$ GDS $\le$ 9), and severe (GDS > 9) based on the depression score \cite{b58}. The distribution of feature data within each subgroup was found to be associated with depression symptoms. The study also revealed that all 9 features exhibited a strong correlation with the symptoms of depression.

\begin{figure*}[htbp]
\vspace{2mm}
\centerline{\includegraphics[width=0.9\linewidth]{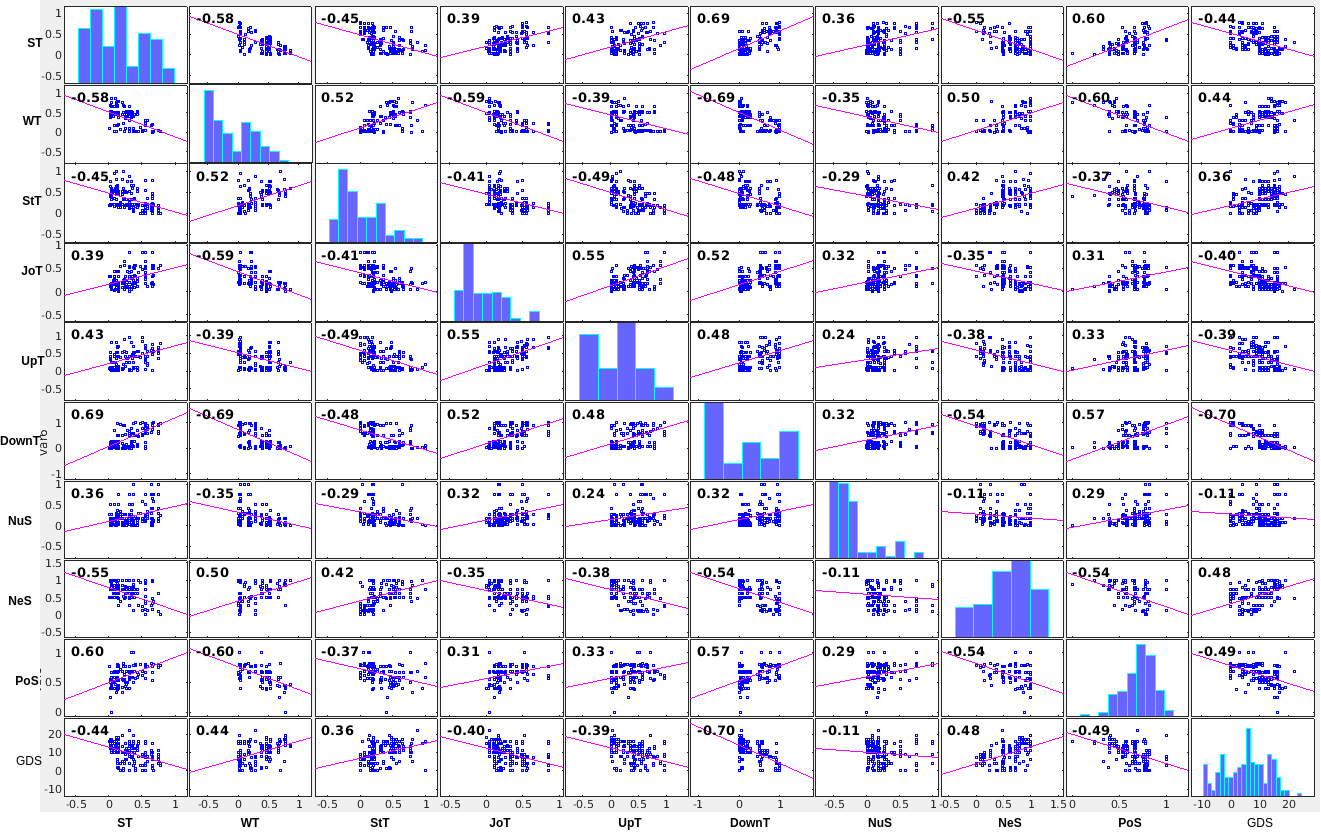}}
\vspace{-3mm}
\caption{\centering{Plot matrix of relationship between all the features, and GDS score. The coefficient of correlation between \newline each activity feature and GDS scores is also shown in the plot matrix.}}
\label{fig 10}
\end{figure*}

\vspace{2mm}
The analysis of the correlation between the features and the GDS scores revealed that 8 of the 9 features were significantly correlated to the scores (Figure \ref{fig 10}). Specifically, sitting, walking, jogging, negative sentiment and positive sentiment showed strong correlations. The $p$ values of sitting, walking, standing, jogging, going upstairs, going downstairs, neutral sentiment, negative sentiment, and positive sentiment are 4.319e-54, 1.072e-43 1.608e-52, 3.081e-50, 9.773e-52, 2.830e-42, 3.305e-34, 9.845e-78, 2.641e-128, 2.969e-67 respectively, which shows the strong relationship of the features. 

\begin{figure}[htbp]
\vspace{-6mm}
\centerline{\includegraphics[scale=0.35]{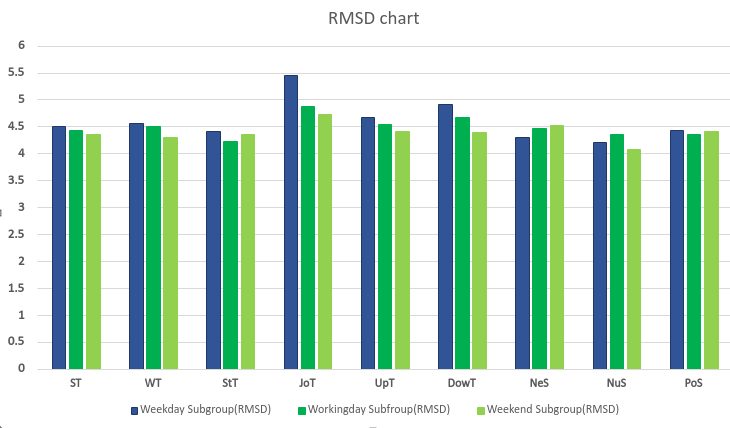}}
\vspace{-4mm}
\caption{Error rate between projected GDS score and genuine score.}
\label{fig 11}
\end{figure}

The connection between the projected GDS score of the regression model and the genuine score appears in figure \ref{fig 11}. 

The correlation between the features was evaluated using RMSD, and lasso regularization was employed to minimize the error rate, as discussed previously. Figure \ref{fig 11} shows the results of each feature calculated using leave-one-out cross-validation. The wrapper feature selection method was used to select a subset of features out of the 27 features in the three subgroups (weekdays, working days, and weekends) that had a lower RMSD of 3.356. This subset indicated stronger connections between the predicted GDS score and the actual GDS score.

\subsection{Depression Level Detection  }

\begin{figure}[htbp]
\vspace{-4mm}
\centerline{\includegraphics[scale=0.27]{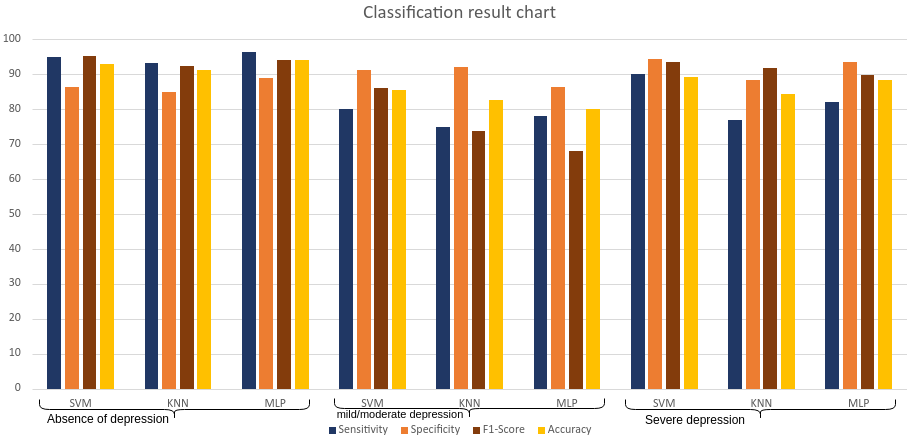}}
\vspace{-2mm}
\caption{Depression level classification results}
\label{fig 12}
\vspace{-4mm}
\end{figure}

Table \ref{table 6a} and Figure \ref{fig 12} present the results of the classification algorithm for the three depression levels (absence, mild/moderate, and severe). From the output in Figure 13, MLP is the top classifier for absence depression, while SVM is the top classifier for mild/moderate and severe depression. The algorithms demonstrate high sensitivity, high specificity, and good F-1 scores, indicating good performance in detecting depression. SVM outperforms the other classification algorithms in terms of accuracy. Although MLP performs better in recognizing the absence of depression, SVM shows satisfactory performance for mild/moderate and severe depression, with high accuracy overall. Therefore, SVM demonstrates better performance for each of the three depression levels. 

\begin{table}[htbp]
    \centering
    \caption{(a) Comparisons between SVM, KNN, and MLP while classifying absence of depression level, (b) Comparisons between SVM, KNN, and MLP while classifying mild/moderator depression level, (c) Comparisons between SVM, KNN, and MLP while classifying severe depression level. }
    \label{table 6a}
    \begin{tabular}{|c|c|c|c|c|}
        \hline
        (A) & Sensitivity & Specificity & F1-Score & Accuracy \\
        \hline
        SVM & 95 & 86.4 & 95.4 & 93 \\
        \hline
        KNN & 93.2 & 84.9 & 92.5 & 91.3 \\
        \hline
        MLP & 96.5 & 89 & 94.2 & 94.2 \\
        \hline
    \end{tabular}
\end{table}

\begin{table}[htbp]
\vspace{-5mm}
    \centering
    \label{table 6b}
    \begin{tabular}{|c|c|c|c|c|}
        \hline
        (B) & Sensitivity & Specificity & F1-Score & Accuracy \\
        \hline
        SVM & 80 & 91.3 & 86.2 & 85.6 \\
        \hline
        KNN & 74.9 & 92.1 & 73.8 & 82.8 \\
        \hline
        MLP & 78 & 86.5 & 68 & 80 \\
        \hline
    \end{tabular}
\end{table}
\begin{table}[htbp]
\vspace{-5mm}
    \centering
    \label{table 6c}
    \begin{tabular}{|c|c|c|c|c|}
        \hline
        (C) & Sensitivity & Specificity & F1-Score & Accuracy \\
        \hline
        SVM & 90 & 94.5 & 93.6 & 89.3 \\
        \hline
        KNN & 77 & 88.3 & 91.8 & 84.5 \\
        \hline
        MLP & 82 & 93.6 & 89.9 & 88.4 \\
        \hline
    \end{tabular}
    \vspace{-2mm}
\end{table}

The ROC (receiver operating characteristic) curves are used to illustrate the execution of the SVM classifier. 
\begin{figure}[htbp]
\vspace{-6mm}
\centerline{\includegraphics[width=0.8\linewidth]{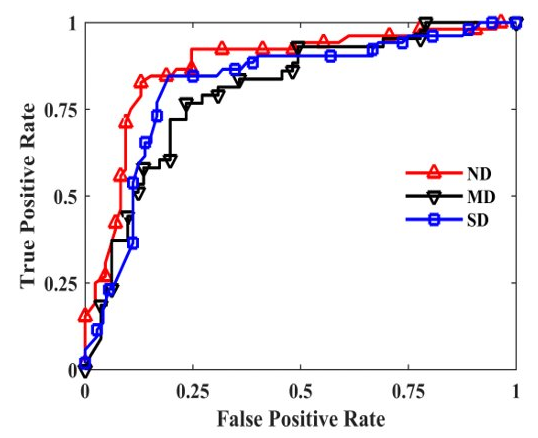}}
\vspace{-4mm}
\caption{TPR vs FPR ROC curve }
\label{fig 13}
\end{figure}

The receiver operating characteristic (ROC) curve illustrates the classification of each and every depression level. As appeared in figure \ref{fig 13}, by taking into consideration the “absence of depression/ no depression (ND)” as an illustrative example of negative test data, the receiver operating characteristic (ROC) bend regarding Depression level arrived at the highest TPR (true positive rate) of around: (I) 82\% for moderate/mild depression (MD),  (ii) 84.5\% for severe depression (SD).

\section{Discussion}
In this comprehensive study, we delve into the intricate world of mental health assessment, employing a multifaceted approach that integrates sentiment analysis from Twitter data, activity recognition, GDS (Geriatric Depression Scale), and depression level detection. Our aim is to unravel the complex interplay of online expressions, physical activities, and participant-specific features in the context of mental well-being. 

The statistical analysis and visualization revealed a notable prevalence of depressed users across different polarity domains, demonstrating the potential of social media as an expressive outlet for mental health concerns. The robust performance of the Naïve Bayes classification algorithm in categorizing tweets into positive, neutral, and negative sentiments underscores the viability of machine learning techniques in extracting valuable insights from unstructured textual data. The observed accuracy of 95.6\% attests to the effectiveness of sentiment analysis in discerning emotional states expressed online.\\
The application of a Convolutional Neural Network (CNN) with depth-wise convolution for activity recognition demonstrated a high accuracy of approximately 98\%. This level of precision in identifying and classifying various activities speaks to the potential of machine learning models in capturing patterns within time-series data. The consistent performance across multiple tests further strengthens the reliability of the CNN model. This underscores the feasibility of utilizing sensor data to gain insights into individuals' physical activities, providing an additional dimension to the holistic understanding of mental well-being.

To improve the model's performance, the "leave-one-out cross-validation" method was applied, where a single participant's features were used for testing while the rest of the participants' features were used for training. The model demonstrated the ability to accurately distinguish between three individual depression levels (absence of depression, mild/moderate depression, and severe depression) with an accuracy of 89.2\%.The regression model employed for predicting GDS scores based on participant activities showcased strong correlations, particularly in activities such as sitting, walking, jogging, and sentiment expression. The leave-one-out cross-validation method ensured a robust evaluation, and the feature selection process emphasized the identification of key variables contributing to accurate predictions. The study found that long periods of sitting and sleeping, less jogging, and negative tweet sentiment were strongly correlated with depression levels. The model was able to identify different levels of depression in participants with an accuracy rate of 94\%, which outperformed previous studies that recorded accuracy rates of 86.5\%\cite{b49}.  The nuanced analysis of GDS scores within subgroups based on participant activities adds granularity to our understanding of the intricate relationship between behavior patterns and depression symptoms.

In this research, the wrapper method was used to improve the accuracy of the model for classifying depression levels (absence of depression, mild/moderate depression, and severe depression). It is challenging to distinguish between mild/moderate depression and the other two subgroups (absence of depression and severe depression). The classification of depression levels using SVM, KNN, and MLP algorithms presented distinct strengths across absence, mild/moderate, and severe depression categories. SVM (85.6\%) emerged as the top  performer for mild/moderate and severe depression, while MLP (94.2\%) exhibited superior performance in detecting the absence of depression. The high sensitivity, specificity, and F1-scores across the three depression levels underscore the potential of machine learning algorithms in providing accurate classifications. The ROC curve analysis further visually encapsulates the discriminatory power of the SVM classifier across different depression levels.

While the current research achieved high accuracy in classifying depression levels, there are several limitations that need to be taken into consideration when interpreting the results. Firstly, the study did not evaluate a clinically depressed population, as the sample consisted only of students, and the depression assessment was based on a self-report scale (GDS) rather than a clinical evaluation. Additionally, there were inconsistencies in the data collection process, as some participants may have forgotten to carry their smartphones, leading to incomplete data. An extended long-term study can be performed in the future with data from clinical side and also for different demographies. Also, words such as "no," "not," "never," "none," and "neither" were deemed overly broad and were intentionally omitted from the category of negative terms in order to enhance precision and specificity in language usage. Finally, privacy policies in social media are becoming increasingly strict, making it difficult to collect data without the user's explicit consent.

\section{Conclusion}
The holistic integration of sentiment analysis, activity recognition, and GDS score prediction into a comprehensive model for depression level detection presents a novel and multidimensional approach. The study not only contributes valuable insights into the understanding of mental health through diverse data modalities but also highlights the potential for technology-driven assessments in clinical and research settings. However, the external validity of the findings requires consideration, and further exploration in diverse populations and contexts is warranted. The multidisciplinary nature of this study positions it at the intersection of data science and mental health research, paving the way for future investigations into technology-enabled mental health diagnostics and interventions. Moreover, including a Smartwatch, wristband, and Fitness Tracker gadget in future examinations would go about as an extra outside wellspring of information assortment to that of the cell phone and would prompt a significantly more elevated level of Data Accuracy. Our future work will focus on optimized model-based neural network compression \cite{b59} for fast and enegy-efficient execution of the framework. 

\section*{Acknowledgment}
 All the participants were informed about the purpose, nature, and procedure of the study, and  were also informed that they have full rights to withdraw their data at any time. After disclosing all the details, formal informed consent was obtained from all the participants. Data privacy was preserved for all the participants. 

\balance


\begin{thebibliography}{00}
\bibitem{b1} World Health Organization, Mental Health Action Plan 2013-2020, http://apps.who.int/iris/bitstream/10665/89966/1/9789241506021\\\_eng.pdf? 
ua=1 [accessed 2018-04-25] 

\bibitem{b2} H.A. Whiteford, L. Degenhardt, J. Rehm, A.J. Baxter, A.J. Ferrari, H.E. Erskine, et al. Global burden of disease attributable to mental and substance use disorders: findings from the global burden of disease study 2010, Lancet;382(9904):1575-1586. 
\bibitem{b3} A. Wongkoblap, M.A. Vadillo, V. Curcin. Researching mental health disorders in the era of social media: systematic review, J Med Internet Res;19(6): e228.
\bibitem{b4} World Health Organization. 2018. Depression: Key Facts   URL: https://www.who.int/news-room/fact-sheets/detail/depression. 
\bibitem{b5} World Health Organization, Depression: Let’s talk, SEARO. (2017). http://www.searo.who.int/bangladesh/enbanwhd2017/en/. 
\bibitem{b6} R.C. Kessler, E.J. Bromet. The epidemiology of depression across cultures Annu. Rev. Public Health, 34 (2013), pp. 119-138.
\bibitem{b7} S.M.Y. Arafat, H. Akter, B. Mali. Psychiatric morbidities and risk factors of suicidal ideation among patients attending for psychiatric services at a tertiary teaching hospital in Bangladesh, Asian J. Psychiatr., 34 (2018), pp. 44-46.
\bibitem{b8} I. Elkin, M.T. Shea, J.T. Watkins, S.D. Imber, S.M. Sotsky, J.F. Collins, D.R. Glass, P.A. Pilkonis, W.R. Leber, J.P. Docherty. National Institute of Mental Health treatment of depression collaborative research program: General effectiveness of treatments Arch. Gen. Psychiatry, 46 (1989), pp. 971-982.
\bibitem{b9} A.J. Mitchell, A. Vaze, S. Rao. Clinical diagnosis of depression in primary care: a meta-analysis, Lancet (2009), pp. 609-619.
\bibitem{b10} Mohammed T. Masud, Mohammed A. Mamun, K. Thapa, D.H. Lee, Mark D. Griffiths, S.-H. Yang. Unobtrusive monitoring of behavior and movement patterns to detect clinical depression severity level via smartphone. Journal of Biomedical Informatics, Volume 103, 2020.
\bibitem{b11} M. Cepoiu, J. McCusker, M.G. Cole, M. Sewitch, E. Belzile, A. Ciampi Recognition of depression by non-psychiatric physicians-a systematic literature review and meta-analysis, J. Gen. Intern. Med., 23 (2008), pp. 25-36. 

\bibitem{b12} R.S. Murthy, K.V.K. Kumar, D. Chisholm, T. Thomas, K. Sekar, C.R. Chandrashekar Community outreach for untreated schizophrenia in rural India: a follow-up study of symptoms, disability, family burden and costs Psychol. Med., 35 (2005), pp. 341-351.
\bibitem{b13} R.C. Kessler, P.A. Berglund, M.L. Bruce, J.R. Koch, E.M. Laska, P.J. Leaf, R.W. Manderscheid, R.A. Rosenheck, E.E. Walters, P.S. Wang. The prevalence and correlates of untreated serious mental illness, Health Serv. Res., 36 (2001), pp. 987-1007. 
\bibitem{b14} Statista, Number of smartphone mobile network subscriptions worldwide from 2016 to 2022, with forecasts from 2023 to 2028, (2023). https://www.statista.com/statistics/330695/number-of-smartphone-users-worldwide/. (accessed 2023). 

\bibitem{b15} Newzoo, Top countries/markets by smartphone penetration \& users, (2018). https://newzoo.com/insights/rankings/top-50-countries-by-smartphone-penetration-and-users/ (accessed 2019). 
\bibitem{b16} A. Grünerbl, A. Muaremi, V. Osmani, G. Bahle, S. Öhler, G. Tröster, O. Mayora, C. Haring, P. Lukowicz. Smartphone-based recognition of states and state changes in bipolar disorder patients, IEEE J. Biomed. Heal. Informatics., 19 (2015).

\bibitem{b17} G.M. Harari, N.D. Lane, R. Wang, B.S. Crosier, A.T. Campbell, S.D. Gosling. Using smartphones to collect behavioral data in psychological science: Opportunities, practical considerations, and challenges, Perspect. Psychol. Sci., 11 (2016).

\bibitem{b18} S. Abdullah, T. Choudhury. Sensing technologies for monitoring serious mental illnesses, IEEE Multimedia, 25 (2018), pp. 61-75. 

\bibitem{b19} D.C. Mohr, M. Zhang, S.M. Schueller. Personal sensing: understanding mental health using ubiquitous sensors and machine learning, Annu. Rev. Clin. Psychol., 13 (2017), pp. 23-47.

\bibitem{b20} E. Garcia-Ceja, V. Osmani, O. Mayora. Automatic stress detection in working environments from smartphones’ accelerometer data: a first step, IEEE J. Biomed. Heal. Informatics., 20 (2016), pp. 1053-1060.

\bibitem{b21} R. Wang, A. Dasilva, J.F. Huckins, W.M. Kelley, T.F. Heatherton, A.T. Campbell, W. Wang, Todd, F. Heatherton. Tracking depression dynamics in college students using mobile phone and wearable sensing, Proc. ACM Interact. Mob. Wearable Ubiquitous Technol. 2, 1(2018) 1–26. 

\bibitem{b22} L.L. Craft, F.M. Perna. The benefits of exercise for the clinically depressed, Prim Care Companion J. Clin. Psychiatry., 6 (2004), pp. 104-111. 

\bibitem{b23} V. Osmani. Smartphones in mental health: detecting depressive and manic episodes, IEEE Pervasive Comput., 14 (2015), pp. 10-13.

\bibitem{b24}S. Saeb, M. Zhang, C.J. Karr, S.M. Schueller, M.E. Corden, K.P. Kording, D.C. Mohr. Mobile phone sensor correlates of depressive symptom severity in daily-life behavior: an exploratory study, J. Med. Internet Res., 17 (2015), p. e175.

\bibitem{b25}L.K. George, D.G. Blazer, D.C. Hughes, N. Fowler. Social support and the outcome of major depression, Br. J. Psychiatry, 154 (1989), pp. 478-485. 

\bibitem{b26}J.Y. Kim, N. Liu, H.X. Tan, C.H. Chu. Unobtrusive monitoring to detect depression for elderly with chronic illnesses, IEEE Sens. J., 17 (2017), pp. 5694-5704.

\bibitem{b27}J.A. Andrews, A.J. Astell, L.J.E. Brown, R.F. Harrison, M.S. Hawley. Technology for early detection of depression and anxiety in older people, Stud. Health Technol. Inform., 242 (2017), pp. 374-380.

\bibitem{b28}T. Nguyen, B. O’Dea, M. Larsen, D. Phung, S. Venkatesh, H. Christensen. Using linguistic and topic analysis to classify sub-groups of online depression communities, Multimed Tools Appl 2015.

\bibitem{b29}P.A. Cavazos-Rehg, M.J. Krauss, S. Sowles, S. Connolly, C. Rosas, M. Bharadwaj, L.J. Bierut. A content analysis of depression-related Tweets. Comput Human Behav. 2016;54:351-357,PMCID: PMC4574287. 
\bibitem{b30}Glen Coppersmith, Mark Dredze, Craig Harman, and Kristy Hollingshead. 2015. From ADHD to SAD: Analyzing the Language of Mental Health on Twitter through Self-Reported Diagnoses. In Proceedings of the 2nd Workshop on Computational Linguistics and Clinical Psychology: From Linguistic Signal to Clinical Reality, pages 1–10. Association for Computational Linguistics  

\bibitem{b31}Glen Coppersmith, Mark Dredze, and Craig Harman. 2014. Quantifying Mental Health Signals in Twitter. In Proceedings of the Workshop on Computational Linguistics and Clinical Psychology: From Linguistic Signal to Clinical Reality, pages 51–60, Baltimore, Maryland, USA. Association for Computational Linguistics.  

\bibitem{b32}J. Radzikowski, A. Stefanidis, K.H. Jacobsen, A. Croitoru, A. Crooks, P.L. Delamater. The measles vaccination narrative in Twitter: a quantitative analysis, JMIR Public Health Surveill 2016;2(1):e1. 

\bibitem{b33}A. Stefanidis, E. Vraga, G. Lamprianidis, J. Radzikowski, P.L. Delamater, K.H. Jacobsen, D. Pfoser, A. Croitoru, A. Crooks. Zika in Twitter: Temporal Variations of Locations, Actors, and Concepts, JMIR Public Health Surveill. 2017 Apr 20. 

\bibitem{b34}K.C. Finch, K.R. Snook, C.H. Duke, K. Fu, Z.T. Tse, A. Adhikari, et al. Public health implications of social media use during natural disasters, environmental disasters, and other environmental concerns. Nat Hazards 2016 Apr 19;83(1):729-760.

\bibitem{b35}G. Eysenbach. Infodemiology and infoveillance: framework for an emerging set of public health informatics methods to analyze search, communication and publication behavior on the internet. J Med Internet Res 2009 Mar 27;11(1):e11. 

\bibitem{b36}M.J. Paul, M. Dredze. You Are What You Tweet: Analyzing Twitter for Public Health, In: Proceedings of the Fifth International AAAI Conference on Weblogs and Social Media. 2011 Presented at: AAAI'11; July 17-21, 2011; Barcelona, Spain p. 265-272. 

\bibitem{b37}M. Park, C. Cha, M. Cha. Depressive Moods of Users Portrayed in Twitter, In: Proceedings of the ACM SIGKDD Workshop on Health Informatics. 2012 

\bibitem{b38}M. Conway, D. O'Connor. Social media, big data, and mental health: current advances and ethical implications. Curr Opin Psychol 2016;9:77-82. 

\bibitem{b39}S.H. Jain, B.W. Powers, J.B. Hawkins, J.S. Brownstein. The digital phenotype. Nat Biotechnol 2015;33(5):462-463. 

\bibitem{b40}Statista. 2018. Number of Monthly Active Twitter Users Worldwide From 1st Quarter 2010 to 2nd Quarter 2018 (in Millions). Y.R. Tausczik, J.W. Pennebaker. The psychological meaning of words: LIWC and computerized text analysis methods, J Lang Soc Psychol 2009;29(1):24-54.

\bibitem{b41}Y.R. Tausczik, J.W. Pennebaker. The psychological meaning of words: LIWC and computerized text analysis methods, J Lang Soc Psychol 2009;29(1):24-54. 

\bibitem{b42}N. mirez-Esparza, C.K. Chung, E. Kacewicz, J.W. Pennebaker. The Psychology of Word Use in Depression Forums in Englishin Spanish: Texting Two Text Analytic Approaches, In: Proceedings of the 2nd International Conference on Weblogs and Social Media. 2008 

\bibitem{b43}V.M. Prieto, S. Matos, M. Álvarez, F. Cacheda, J.L. Oliveira. Twitter: a good place to detect health conditions, PLoS One 2014;9(1):e86191. 

\bibitem{b44}R. Thackeray, S.H. Burton, C. Giraud-Carrier, S. Rollins, C.R. Draper. Using Twitter for breast cancer prevention: an analysis of breast cancer awareness month, BMC Cancer 2013;13:508.

\bibitem{b45}Sherry A. Greenberg, The Geriatric Depression Scale (GDS), Hartford Institute for Geriatric Nursing, NYU College of Nursing, https://wwwoundcare.ca/Uploads/ContentDocuments/Geriatric\%20 Depression\%20Scale.pdf. 

\bibitem{b46}Mia Conradsson, Erik Rosendahl, Håkan Littbrand, Yngve Gustafson, Birgitta Olofsson, Hugo Lövheima. Usefulness of the Geriatric Depression Scale 15-item version among very old people with and without cognitive impairment, Aging Ment Health. July, 2013.

\bibitem{b47}Python-Twitter API, Getting your application tokens, https://python-twitter.readthedocs.io/en/latest/getting\_started.html. 

\bibitem{b48}D.M. Karantonis, M.R. Narayanan, M. Mathie, N.H. Lovell, B.G. Celler Implementation of a real-time human movement classifier using a triaxial accelerometer for ambulatory monitoring, IEEE Trans. Inf. Technol. Biomed., 10 (2006).

\bibitem{b49}S. Saeb, M. Zhang, C.J. Karr, S.M. Schueller, M.E. Corden, K.P. Kording, D.C.Mohr. Mobile phone sensor correlates of depressive symptom severity in daily-life behavior: an exploratory study, J. Med. Internet Res., 17 (2015), p. e175, 10.2196/jmir.4273. 

\bibitem{b50}Naive Bayes machine learning algorithm, scikit-learn ,https://scikit-learn.org/stable/modules/naive\_bayes.html.  

\bibitem{b51}S.-H. Yang, M.H. Kabir, M.R. Hoque. Mathematical modeling of smart space for context-aware system: linear algebraic representation of state-space method based approach, Math. Probl. Eng., 2016.

\bibitem{b52}R. Tibshirani. Regression shrinkage and selection via the lasso:a retrospective, J. R. Stat. Soc. Ser. B., 73 (1996).

\bibitem{b53}I. H. Witten, E. Frank, ed. Data Mining: Practical Machine Learning Tools and Techniques, Fourth ed., Los Altos: Morgan Kaufmann, San Francisco, 2017. 

\bibitem{b54}D.W. Aha, D. Kibler, M.K. Albert. Instance-based learning algorithms, Mach. Learn., 6 (1991), pp. 37-66.

\bibitem{b55}Bayat, F.M. Prezioso, M.B. Chakrabarti, et al. Implementation of multilayer perceptron network with highly uniform passive memristive crossbar circuits, Nat Commun 9, 2331 (2018).

\bibitem{b56}R. Kohavi, G.H. John. Wrappers for feature subset selection, Artif. Intell., 97 (1997), pp. 273-324.

\bibitem{b57}T. Tekin Erguzel, C. Tas, M. Cebi. A wrapper-based approach for feature selection and classification of major depressive disorder-bipolar disorders Comput., Biol. Med. 64 (2015).

\bibitem{b58}L.E. Stone, K.L. Granier, D.L. Segal. (2019). Geriatric Depression Scale, In: Gu D., Dupre M. (eds) Encyclopedia of Gerontology and Population Aging. Springer, Cham. 

\bibitem{b59}M. H. Uddin, J. M. K. Ara, M. H. Rahman and S. H. Yang, "Neural network pruning: An effective way to reduce the initial network for deep learning based human activity recognition," 2021 International Conference on Electronics, Communications and Information Technology (ICECIT), 2021.

\end{thebibliography}
\end{document}